\documentclass[11pt]{article}
\usepackage{acl}

\usepackage{times}
\usepackage{latexsym}
\usepackage{todonotes}
\usepackage{amsmath}
\usepackage{multirow}
\usepackage{bbm}
\usepackage{caption}
\usepackage{subcaption}
\usepackage[T1]{fontenc}
\usepackage[utf8]{inputenc}
\usepackage{microtype}
\usepackage{comment}
\usepackage{graphicx}

\newcommand{\para}[1]{\smallskip\noindent\textbf{#1.}}

\newcommand*{\probe}{\texttt{\sc ConPare-LAMA}}
\AtBeginDocument{%
  \providecommand\BibTeX{{%
    \normalfont B\kern-0.5em{\scshape i\kern-0.25em b}\kern-0.8em\TeX}}}

\begin{document}

\title{Dissecting Paraphrases: \\ The Impact of Prompt Syntax and Supplementary Information on Knowledge Retrieval from Pretrained Language Models}
\author{\\ Stephan Linzbach $\diamondsuit\spadesuit$,
        Dimitar Dimitrov$\diamondsuit$, 
         Laura Kallmeyer$\spadesuit$, \\
         Kilian Evang$\spadesuit$, 
         Hajira Jabeen$\diamondsuit$\thanks{\ \ \ Corresponding Author}, \and
         Stefan Dietze$\diamondsuit\spadesuit$* \\
         $\diamondsuit$ GESIS - Leibniz Institute for the Social Sciences, name.surname@gesis.org \\
         $\spadesuit$ Heinrich Heine University, name.surname@hhu.de \\
         }
 
\maketitle
\begin{abstract}
  Pre-trained Language Models (PLMs)
  are known to contain various kinds of knowledge. One method to infer relational knowledge is through the use of cloze-style prompts, where a model is tasked to predict missing subjects or objects.
  Typically, designing these prompts is a tedious task because small differences in syntax or semantics can have a substantial impact on knowledge retrieval performance. 
  Simultaneously, evaluating the impact of either prompt syntax or information is challenging due to their interdependence. 
  We designed \probe{} -- a dedicated probe, consisting of 34 million distinct prompts that facilitate comparison across minimal paraphrases.
  These paraphrases follow a unified meta-template enabling the controlled variation of syntax and semantics across arbitrary relations.
  \probe{} enables insights into the independent impact of either syntactical form or semantic information of paraphrases on the knowledge retrieval performance of PLMs. 
  Extensive knowledge retrieval experiments using our probe reveal that prompts following clausal syntax have several desirable properties in comparison to appositive syntax: i) they are more useful when querying PLMs with a combination of supplementary information, ii) knowledge is more consistently recalled across different combinations of supplementary information, and iii) they decrease response uncertainty when retrieving known facts. In addition, range information can boost knowledge retrieval performance more than domain information, even though domain information is more reliably helpful across syntactic forms.
\end{abstract}

\section{Introduction}
\label{sec:introduction}

Symbolic knowledge bases provide relational knowledge and are widely used for tasks like question-answering. 
However, they rely on costly manual or automated, often supervised, information extraction pipelines to retrieve and represent relational knowledge. 
Relational knowledge refers to knowledge about relations between entities, e.g. `Paris', `capitalOf', `France', where `capitalOf' is the \emph{relation}, `Paris' is the \emph{subject}, and `France' is the \emph{object}. 
Previous research on relational knowledge retrieval (\textit{rKR}) from pre-trained language models (PLMs) \cite{petroni-etal-2019-language, sung2021can} has demonstrated that relational knowledge can be retrieved directly from the parameters of a PLM. 
This finding has led to a plethora of research concerned with knowledge retrieval and reasoning capacities of PLMs \cite{petroni-etal-2019-language, sung2021can, zhong2021factual, elazar2021measuring,jiang2019know}, where \textit{rKR} performance is seen as an indicator of PLM's capacities to understand and reason.
Several benchmarks have been proposed that aim at measuring \textit{rKR} performance as the ability of a PLM to predict masked 
objects as part of cloze-style prompts \cite{petroni-etal-2019-language, kalo2022kamel, kassner2021multilingual}.
It was found that some types of supplementary information (sInf) 
are helpful to PLMs \cite{cao2021knowledgeable, petroni2020context, chen2022knowprompt} while other types deteriorate knowledge retrieval performance \cite{pandia2021sorting, kassner-schutze-2020-negated}, and that PLMs primarily rely on memorization, hence, low-frequency examples are less well remembered \cite{ravichander2020systematicity}. 
Additionally, prior works have shown that \textit{rKR} through prompts is inconsistent across different paraphrases \cite{elazar2021measuring, heinzerling2020language}.

Paraphrasing a prompt may introduce a variety of changes, including \textbf{semantic} ones that change the information content of the prompt, i.e., \emph{domain} information (`Paris is a city and is the capital of [MASK]') or \emph{range} information (`Paris is the capital of [MASK], which is a country.'), as well as \textbf{syntactic} ones that merely change the form in which the same content is expressed i.e., \emph{clausal} (`Paris is a city and is the capital of [MASK]') or an \emph{appositive} syntax (`The city Paris is the capital of [MASK]').

 \begin{figure*}[h!]
    \centering
    \includegraphics[width=\textwidth]{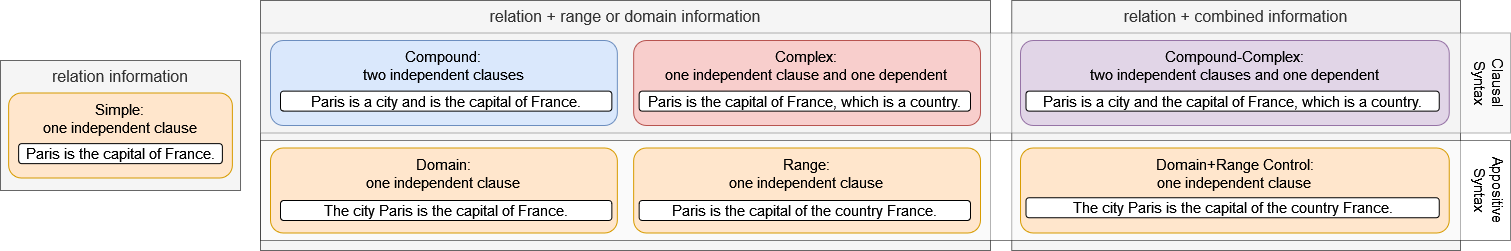}
    \caption{Relationship between prompt types (simple (orange), compound (blue), complex (red), compound-complex (purple)), syntactic forms (clausal and appositive), and sInf combinations (relation, relation+range or domain, relation+combined) used to study the influence of syntax on knowledge retrieval.}
    \label{fig:prompt_data}
    \vspace{-1em}
\end{figure*}

Previous works have not evaluated the combined impact of syntax and semantics or struggled to control all involved variables \cite{linzbach2023decoding, elazar2021measuring, heinzerling2020language}. 

Thus, dedicated probes are required that can control the effects of different syntactic and semantic realisations on \textit{rKR}. 

Our main contributions include: 

\textbf{Controlled Paraphrasing.} 
We apply a meta-template that streamlines prompt engineering across arbitrary relations while enabling control over syntactic form and semantic content (\S\ref{sec:sapd}). 
In terms of syntactic form, in English single-sentence prompts, information must be added either in the form of a noun phrase appositive or as an additional clause. 
The meta-template covers prompts with clausal (as compound, complex, or compound-complex) and appositive syntax. 
Additionally, it includes placeholders for sInf as domain or range information or both. 
In our case, automated prompt construction is enabled by fetching domain/range information (as sInf) for given relations from established knowledge bases such as Wikidata. 
Given this method, we can compare paraphrases focused on particular combinations of semantics and syntax while controlling for variables not under investigation. 
\textbf{Probe and benchmark (\probe{}).} We introduce \probe{}\footnote{Supplementary materials and source code submitted.}, a novel \textbf{Con}trolled \textbf{Par}aphrasing Prob\textbf{e} for \textbf{LAMA} (\S\ref{sec:probe}), that is to the best of our knowledge the first \textit{rKR} probe that facilitates extensive experiments by controlling for both syntax and semantics of prompts and is the largest \textit{rKR} probe so far publicly released. We investigate a set of 60 relations, derived from the established LAMA-probe, ensuring wide comparability with other knowledge retrieval research. More specifically, we utilise the TREx, GoogleRE and ConceptNet corpora from LAMA as described in \citet{petroni-etal-2019-language}. For each relation, using our meta-template, we generate prompts where sInf is added to the subject, the object, both, or neither. The sInf is realized with different syntax, resulting in a total of seven prompts per relation.
Varying the sInf obtained from Wikidata for each such prompts results in roughly 34 million prompts (TREx: 7 mio, ConceptNet: 26 mio, GoogleRE: 1 mio) unique prompts contained in \probe{}. 
\textbf{Experiments and findings.} We conduct experiments on the base versions of three well-established PLMs, i.e., BERT \cite{devlin2018bert}, RoBERTa \cite{liu2019roberta} and Luke \cite{yamada2020luke} to advance the understanding of \textit{rKR} from PLMs. 
In particular, we investigate the following research questions. 
[RQ 1:] What is the impact of prompt syntax and information on \textit{rKR} performance? [RQ 2:] What is the impact of syntax on the PLMs' ability to efficiently combine sInf in prompts?; [RQ 3:] How consistent are the answers of PLMs when comparing the set of correctly retrieved samples for different prompt syntax and information content?; [RQ 4:] How does prompt syntax impact the response uncertainty of PLMs? We find that all models perform better on prompts using sInf through clausal syntax, on all investigated corpora, as compared to information added via appositives. BERT achieves the best performance in this case. In addition, knowledge retrieved through prompts that rely on clausal syntax when adding sInf is more consistent given assumptions about the a priori knowledge available in the prompt. Underlining these findings is our observation that the uncertainty of models for responses to \emph{known facts} decreases when adding sInf through clausal syntax, which is not the case for appositive syntax.

\section{Related Work}
\label{sec:related}
In this section, we provide the necessary background and discuss
prior work.

\para{Knowledge in PLMs}
Since the proposal of transformer-based PLMs \cite{devlin2018bert}, huge efforts have been spent to analyse the knowledge encoded in the learned representations \cite{rogers2021primer}.
All conceivable types of knowledge are tested.
(i) Syntactic and general linguistic knowledge \cite{ettinger2020bert, hewitt2019structural, liu2019linguistic, htut2019attention, goldberg2019assessing, tenney2019you, clark2019does}, where \citet{swayamdipta2019shallow} show that PLMs do not benefit from shallow syntactical features. \citet{reif2019visualizing} show correlations between the gold standard dependency tree and the attention mechanisms in PLMs. \citet{hewitt2019structural} show for BERT \citep{devlin2018bert} that syntax trees are consistently embedded by the neural network.
(ii) Performance in knowledge driven tasks \cite{bosselut-etal-2019-comet, radford2019language, da2019cracking, talmor2020olmpics, warstadt2019investigating, sung2021can, gao2022kmir}.
(iii) Investigation of relational knowledge inherent in PLMs through the LAMA-probe \cite{petroni-etal-2019-language}.
This probe contains knowledge-targeted cloze-style prompts for querying of different models. 
The idea, behind the LAMA-probe motivated researchers to investigate multilingual knowledge \cite{kassner2021multilingual}, knowledge about entities with more complex naming \cite{kalo2022kamel}, and the impact of prompt phrasing on retrieval performance \cite{jiang2020can}.

\para{Consistency of PLMs}
Research concerned with the consistency analyses the answer space of PLMs queried for the same fact through various cloze-style prompts \cite{heinzerling2020language, elazar2021measuring}.
Testing the consistency of PLMs regarding negation and mispriming \citet{kassner-schutze-2020-negated} showed that PLMs are mostly insensitive to the notion of negation and distracted by mispriming.
The latter finding was additionally strengthened by \citet{pandia2021sorting}, \citet{misra2020exploring}, and lately confirmed for LLMs \cite{shi2023large} where irrelevant context was used to distract `code-davinci-002' from the GPT3 family.
In comparison to \citet{jiang2020can} that used paraphrases to investigate peak knowledge, and research that investigated semantic perturbation \cite{kassner-schutze-2020-negated, misra2020exploring, pandia2021sorting}, \citet{elazar2021measuring} introduced the PARAREL probe with which they investigate consistency of language models across prompt paraphrases. They conclude that the models have a generally low consistency. 
Considering, the justified distrust in PLM \textit{rKR} performance \citet{petroni2020context} and \citet{cao2021knowledgeable} try to understand the impact of helpful information in prompts. 
They find that a wide array of sInf helps the models to increase retrieval performance.
Thus, motivating the research in the field of prompt engineering \citet{hu2021knowledgeable}, KnowPrompt \cite{chen2022knowprompt}, and an Ontology based proposal by \citet{ye2022ontology}.

\para{Our research} 
Our work builds upon the ideas proposed by the LAMA-probe. 
However, we study the influence of prompt paraphrases by controlling for syntax and semantic change on \textit{rKR} performance, not the general capacity of PLMs \cite{zhong2021factual, petroni-etal-2019-language}. 
Whereas \citet{cao2021knowledgeable} investigate the impact of an array of sInf on the models' performance, we study how sInf is differently incorporated depending on syntax. 
In contrast to \citet{elazar2021measuring} we measure consistency of PLMs in different syntactic and semantic scenarios.

\begin{figure*}[ht!]
    \centering
    \includegraphics[width=\textwidth]{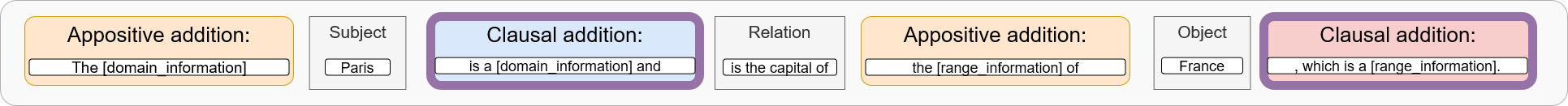}
    \caption{Meta-template that facilitates comparable prompt creation for various relations and information demands.}
    \label{fig:meta_template}
\end{figure*}
   
\begin{table*}[h!]
\small
    \centering
    \begin{tabular}{|c|c|c|}
         \hline
         Relation & LAMA & Ours \\
         \hline
         Occupation & [S] is a [MASK] by profession. & [S] has occupation [MASK]. \\
         Native Language & The native language of [S] is [MASK]. & [S] natively speaks [MASK]. \\
         HasProperty & The [world] today is getting more and more [MASK]. & [S] can be described as [MASK]. \\
         \hline
    \end{tabular}
    \caption{LAMA prompts with high syntactical variation vs. our schematic prompt style.}
    \vspace{-1em}
    \label{tab:relations}
\end{table*}
\section{Controlled Paraphrasing}
\label{sec:sapd}
We propose \probe{} (\textbf{Con}trolled \textbf{Par}arphrasing Prob\textbf{e} for \textbf{LAMA}) to investigate how syntax and semantics of paraphrases impact  knowledge retrieval performance of PLMs. 
We hypothesize that certain syntactic forms facilitate correct interpretation, while certain semantic additions are more useful than others. 
As we are working with relations, we found that a natural addition of information would be domain and range type constraints. 
Using this as sInf, we hardly change the semantics of the original task of \textit{rKR}. 
Furthermore, we restricted our work to single-sentence English prompts. 
We can classify all such sentences as either \textit{clausal}
(cf. Fig.~\ref{fig:prompt_data}, upper row) or \textit{appositive} syntax
(cf. Fig.~\ref{fig:prompt_data}, lower row). 
Additionally, we are interested in the sentence's overall shape, as reflected by the classification of sentences by traditional grammars as \emph{simple} (orange), \emph{compound} (blue), \emph{complex} (red), or \emph{compound-complex} (purple) \citep[cf. ][]{huddleston1984introduction}.
We introduce our probe in three steps: first we \textit{control prompt syntax and semantic effects} (\S\ref{sec:prompttypes}) by selecting prompt sentence types and sInf, then we describe the \textit{meta-template definition} (\S\ref{sec:templateconstruction}), and lastly we show the automatic \textit{template instantiation} (\S\ref{sec:templateinstantiation}). 

\subsection{Dissecting Paraphrasing}%Prompt Types
\label{sec:prompttypes}
Previous work has shown that subtle changes in the phrasing of the prompt can have a substantial impact on model predictions \cite{jiang2020can, elazar2021measuring}. 
However, changes of the semantic information in a prompt are not yet studied independently of their syntactic realisation and vice versa.
We aim to isolate the effects of syntax and semantic change on \textit{rKR}. 
For this, we control the the syntactic realisation of the prompt while varying the sInf. 
As we test several syntactic realisation we can also observe the impact of semantic change.

\para{Prompt sentence types} 
Our prompt construction starts from the \emph{simple} type (orange box in the upper left of Fig.~\ref{fig:prompt_data}). It consists of one main clause that encodes the basic knowledge triple (`Paris is the capital of France'). Contrast this with the \emph{clausal} types that add supplementary domain information in the form of a coordinated clause (blue, `Paris is a city and is the capital of France') or range information in the form of a subordinated clause (red, `Paris is the capital of France, which is a country'), or both (purple, `Paris is a city and is the capital of France, which is a country'). Following \citet{huddleston1984introduction}, we call these clausal prompt types \emph{compound}, \emph{complex}, and \emph{compound-complex}, respectively. 
We used coordination (compound) for subjects and subordination (complex) for objects because these forms were deemed to be the smallest natural-sounding clausal additions that contain the respective domain/range information (i.e., using coodination for object would require duplicating the masked object \emph{Paris is the capital of [MASK] and [MASK] is a country}.
Analogously, prompts with \emph{appositive} syntax are used to add the equivalent sInf (orange boxes in the lower row).

\para{Supplementary Information}
As we intend to assess the slot-filling performance on prompts with syntax going beyond the simple syntax used in LAMA-probe, we use sInf (e.g. city, European country etc.) to construct complex or compound prompt types. 
To fetch this information, we utilise domain and range type constraints for the objects and subjects in respective relations.
For example, in Wikidata the relation `capital' (P:36\footnote{https://www.wikidata.org/wiki/Property:P36}) has the domain type restriction\footnote{https://www.wikidata.org/wiki/Q21503250}: `area', `geographic region', and `fictional planet' etc., and the {range type restriction}\footnote{https://www.wikidata.org/wiki/Q21510865}: `political territorial entity', `fictional city', and `capital city' etc.
This enables us to dynamically generate paraphrases with sInf.

\subsection{Meta-Template Definition}
\label{sec:templateconstruction}
The LAMA-probe tests the assumption that PLMs can function as knowledge bases by manually crafting prompts. These prompts are written to achieve reasonable retrieval performance with no focus on syntactic features. 
Furthermore, prompts for several corpora are 
written to query a single knowledge triple. 
We, however, are interested in \textit{syntactic comparability} between all prompts. Therefore, we introduce a meta-template (cf. Fig. \ref{fig:meta_template}). 
In our meta-template, the grey boxes (i.e., subject, relation, object) are mandatory, while the colored boxes (orange, blue, red) are prompt type specific additions. 
Purple indicates the combination of blue (compound) and red (complex) to form a sentence following the compound-complex typology (analogue for appositive). 
Applying this meta-template avoids confounding effects of relation-specific syntactic forms on retrieval performance. 
We manually crafted a natural-language encoding of each relation that fits the meta-template.
A brief comparison of \probe{} and LAMA-probe can be seen in Tab. \ref{tab:relations}.
Note that the prompt for relation ('HasProperty') is uniquely written for the triple (`world', `HasProperty', `complicated') in the LAMA-probe. 
Moreover, our meta-template assumes that we can use the same template for all triples of one relation and that relation and object text always remain in the same main clause.

\subsection{Template Instantiation}
\label{sec:templateinstantiation}
To instantiate our meta-template, we propose two completion strategies for selecting supplementary (range/domain) information: 

\para{Quality Completion} - choosing the information that leads the model to predict the \textit{right} token with the highest probability. 

\para{Confidence Completion} - choosing the information that leads the model to predict \textit{any} token with the highest probability. 

\subsection{ConPare-LAMA}
\label{sec:probe}
We adapt the LAMA-probe with our controlled probe design to introduce \probe{} (\textbf{Con}trolled \textbf{Par}aphrasing Prob\textbf{e} for \textbf{LAMA}). 
Domain and range type constraints depend on relations. 
Hence, \probe{} contains only triple-based LAMA probe corpora (i.e. TREx, GoogleRE, ConceptNet). 
All corpora are reduced to a comparable size, meaning only triples where the object is in the token vocabulary of all studied models are considered (cf. \probe{} statistics in Tab. \ref{tab:stats_table}). 
We manually expressed all used relations in natural language statements, in a minimal fashion to fit the meta-template. 
For the 41 Wikidata relations available in TREx, we queried the domain\footnote{https://www.wikidata.org/wiki/Q21503250} and range\footnote{https://www.wikidata.org/wiki/Q21510865} type constrains from Wikidata. 
For five relations there is no sInf available. 
However, we manually define one type constraint for the range and the domain for those five relations to ensure that all prompts can be instantiated. 
To get the domain and range information for GoogleRE, we translated the given relations to their Wikidata counter-part. 
For the 16 relations in ConceptNet, we mapped our manually chosen type constraints to their noun concept in ConceptNet.
From those concepts we inferred domain and range type constraints using all concepts connected to the seed concept via `related to' or `defined by' relations.

\begin{table}[t]
    \resizebox{\columnwidth}{!}{
    \centering
    \begin{tabular}{|c|c|c|c|c|c|}
         \hline
         Corpus & Grouping & \#Relations & \#Facts  & Dom & Rng \\
         \hline
         \multirow{4}{*}{TREx} & 1:1 & 2 & 651 & 6.5 & 5.5 \\
         & N:1 & 23 & 18682 & 9.5 & 6.4 \\
         & N:M & 16 & 10190 & 15.6 & 10.1 \\
         & Total & 41 & 29523 & 11.6 & 7.7 \\
         \hline
         \hline
         \multirow{3}{*}{GoogleRE} & death place & 1 & 649 & 10 & 10 \\
         & birth place & 1 & 2404 & 7 & 8 \\
         & birth date & 1 & 1565 & 16 & 1 \\
         & Total & 3 & 4618 & 11 & 6.3 \\
         \hline
         \hline
         ConceptNet & Total & 16 & 22739 & 13.2 & 12.6 \\
         \hline
    \end{tabular}
    }
    \caption{\probe{} corpora statistics with mean number of available object domain (Dom) and range (Rng) types per relation as defined in Wikidata.}
    \label{tab:stats_table}
    %\vspace{-2em}
\end{table}

\section{Experiments}
\label{sec:setup}
\begin{table*}[h!]
\small
\begin{center}
 \tabcolsep=2pt
    \centering
    \begin{tabular}{|c|c|c|c|c|c|c|c|c|c|c|c|c|c|c|c|c|}
        \hline
         \multirow{3}{*}{Corpus} & \multirow{3}{*}{Grouping} & \multicolumn{3}{c}{Relation} &\multicolumn{6}{|c|}{Relation + Domain Information} & \multicolumn{6}{c|}{Relation + Range Information} \\
          & & BERT & 
           RoB & Luke & \multicolumn{2}{c|}{BERT} & \multicolumn{2}{c}{RoBERTa} & \multicolumn{2}{|c|}{Luke} & \multicolumn{2}{c}{BERT} & \multicolumn{2}{|c|}{RoBERTa} & \multicolumn{2}{c|}{Luke} \\
          & & \multicolumn{3}{c|}{Simple} & Cpnd & Appo & Cpnd & Appo & Cpnd & Appo &  Cplx & Appo &  Cplx & Appo &  Cplx & Appo \\
          \hline
         \multirow{4}{*}{TREx} & 1:1 & .4439 & .3118 & .3394 & \underline{.6405} & .5760 & .5238 & \underline{.6098} & .5330 & \underline{.5944} & \underline{.6205} & .6052 & \underline{.5775} & .5238 & \underline{.5668} & .5176 \\
& N:1 & .2876 & .1956 & .2240 & \underline{.3329} & .3086 & \underline{.2706} & .2316 & \underline{.2792} & .2500 & \underline{.3656} & .2919 & \underline{.3547} & .2722 & \underline{.3652} & .2496 \\
& N:M & .2517 & .2205 & .2401 & .3217 & \underline{.3261} & .2986 & \underline{.3166} & \underline{.3168} & .3105 & \underline{.3488} & .1871 & \underline{.2979} & .2121 & \underline{.3015} & .1879 \\
& Total & \textbf{.2786} & .2067 & .2321 & \underline{\textbf{.3358}} & .3205 & \underline{.2859} & .2693 & \underline{.2978} & .2785 & \underline{\textbf{.3654}} & .2627 & \underline{.3400} & .2570 & \underline{.3477} & .2342 \\
          \hline
          \hline
         \multirow{4}{*}{GoogleRE}  & birth-date & .0 & .0012 & .0191 & .0 & .0 & .0178 & \underline{.0364} & .0319 & \underline{.0428} & \underline{.0044} & .0 & \underline{.0083} & .0031 & .0031 & .0031 \\
& birth-place & .1738 & .1156 & .0183 & \underline{.2129} & .1855 & \underline{.0994} & .0715 & \underline{.0590} & .0345 & \underline{.2254} & .2029 & \underline{.1863} & .1730 & \underline{.2104} & .1988 \\
& death-place & .1479 & .0061 & .0015 & \underline{.1479} & .1263 & .0061 & \underline{.0077} & \underline{.0154} & .0077 & .1571 & \underline{.1771} & \underline{.1587} & .1510 & \underline{.1879} & .1448 \\
& Total & \textbf{.1113} & .0614 & .0162 & \underline{\textbf{.1316}} & .1143 & \underline{.0586} & .0506 & \underline{.0437} & .0335 & \underline{\textbf{.1409}} & .1305 & \underline{.1221} & .1123 & \underline{.1370} & .1249 \\
\hline
\hline
ConceptNet & Total & \textbf{.0229} & .0226 & .0230 & \underline{.0455} & .0390 & \underline{\textbf{.0512}} & .0463 & .0494 & \underline{.0507} & \underline{.0495} & .0084 & \underline{\textbf{.0586}} & .0098 & \underline{.0484} & .0088  \\
\hline
\end{tabular}
    \caption{Performance (P@1) when querying with the base typologies and respective appositive. Underline indicates per row and model winner of either the clausal or the appositive prompt. Bold indicates the best performance across all models per corpus and available information.}
    \label{tab:results_one_strategy_one}
    \vspace{-2em}
\end{center}
\end{table*}

We use base models of three different PLMs for evaluation: BERT \cite{devlin2018bert} as it is a well established model and it has already been assessed using the LAMA probe, 
RoBERTa \cite{liu2019roberta} as it is shown to be superior in performance on a range of downstream tasks when compared to BERT,  and Luke \cite{yamada2020luke} as it uses entity word cross-attention to enhance the knowledge of a base RoBERTa model. 
However, to keep it comparable, we only input the tokenized text without entity span information. 
If not mentioned explicitly we complete the prompts that either add domain or range information with \textit{Quality Completion} and reuse this information to populate the prompt that encodes both kinds of information. 
This is done to ensure best possible performance per prompt and triple. 
We use the P@1 metric to measure \textit{rKR} performance following \citet{petroni-etal-2019-language}. 
We only conduct our experiment on base models as 
consistency concerning paraphrases only marginally increases from base to large configurations of the PLMs \citep{elazar2021measuring}. 

\subsection{Results}

\para{Impact of syntax and semantic on knowledge retrieval performance (RQ1)}
\label{ssec:Performance}
We organize the process of paraphrasing across two dimensions, the semantic dimension (i,e., changing sInf), and the syntactic dimension (i.e., changing the sentence type) where we consider \emph{clausal} (compound=Cpnd, complex=Cplx) and \emph{appositive} syntax (Appo). 
Table \ref{tab:results_one_strategy_one} offers three perspectives on the impact of paraphrasing (from top to bottom) (1) observing performance change across semantic paraphrases (Relation, Relation+Domain Information, Relation+Range Information), (2) the impact of paraphrasing per model (BERT, RoBerta, Luke) and (3) performance change for syntactic paraphrasing of semantically equivalent content (Simple, Cpnd, Cplx, Appo). 
We can measure the impact of semantic change by comparing the best performance per semantic category per model. 
We observe that sInf through prompts using \emph{clausal} (compound=Cpnd, complex=Cplx) syntax increases the performance for all three models on all corpora. 
For the \emph{appositive} syntax (Appo), this is generally true, with a few exceptions, though the increase is relatively less across relations than with clausal syntax, except for Luke on ConceptNet.
Furthermore, we can see that the performance given domain information is more stable across different syntaxes when compared to the addition of range information.
However, adding range information has the most potential to increase performance. 
When we compare this across different models we observe a common trend in all of them. 
Lastly, we analyze the impact of syntactic paraphrasing. 
In general \emph{clausal} prompts outperform their \emph{appositive} counterpart. 
In particular for TREx, the average performance gain from sInf is weaker for compound prompts ($\approx 2\%$) than for complex prompts ($\approx 10\%$) when compared to their respective appositive counterparts. 
Expanding upon the findings presented by \citet{petroni2020context} and \citet{cao2021knowledgeable}, we show that supplementing range and domain information is helpful for \textit{rKR}.

\begin{figure*}[h!]
     \centering
     \begin{subfigure}[b]{\columnwidth}
         %\centering
         \includegraphics[width=\columnwidth]{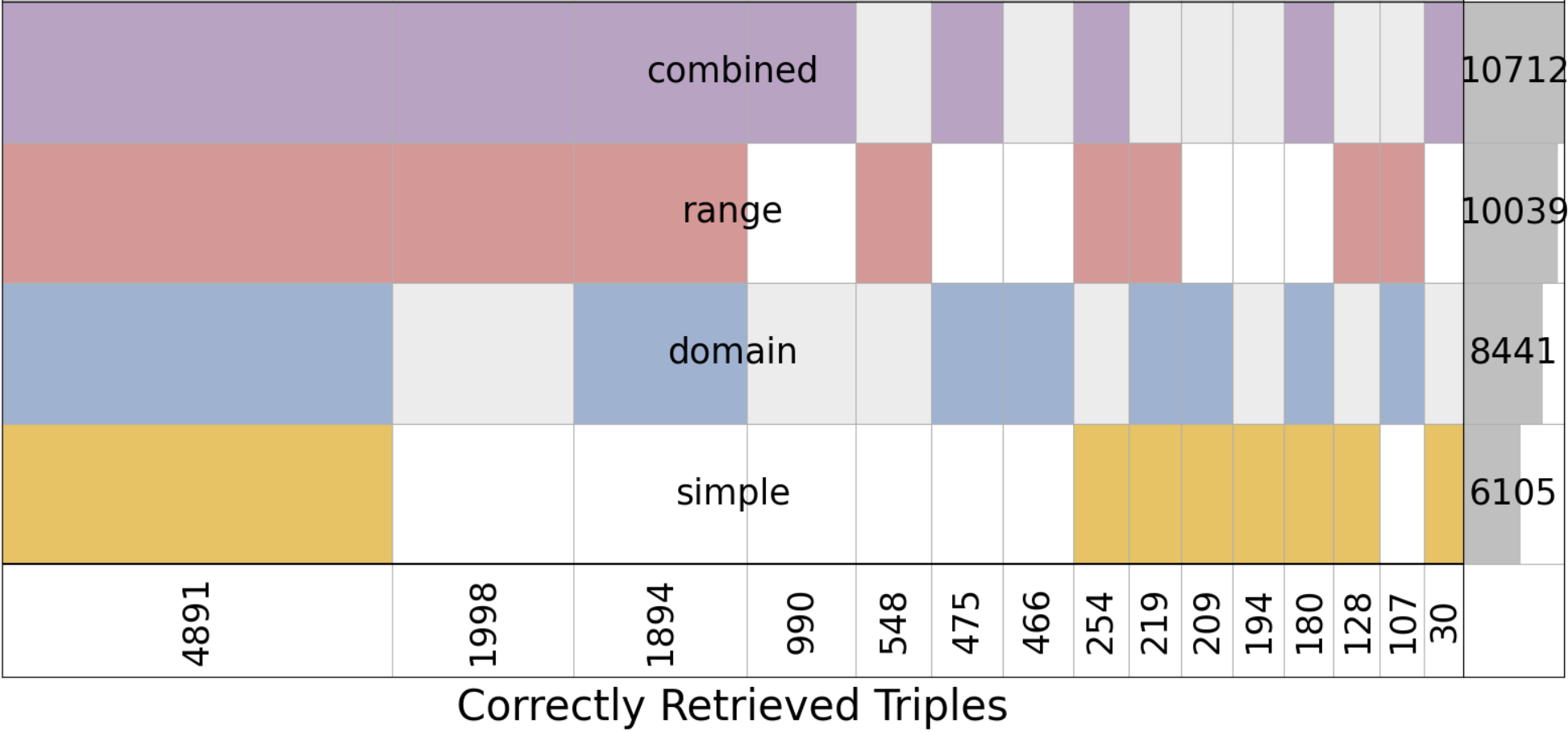}
         %\vspace{-2em}
         \caption{Clausal prompts}
         \label{fig:clausal_consistency}
     \end{subfigure}
     \hfill
     \begin{subfigure}[b]{\columnwidth}
         %\centering
         \includegraphics[width=\columnwidth]{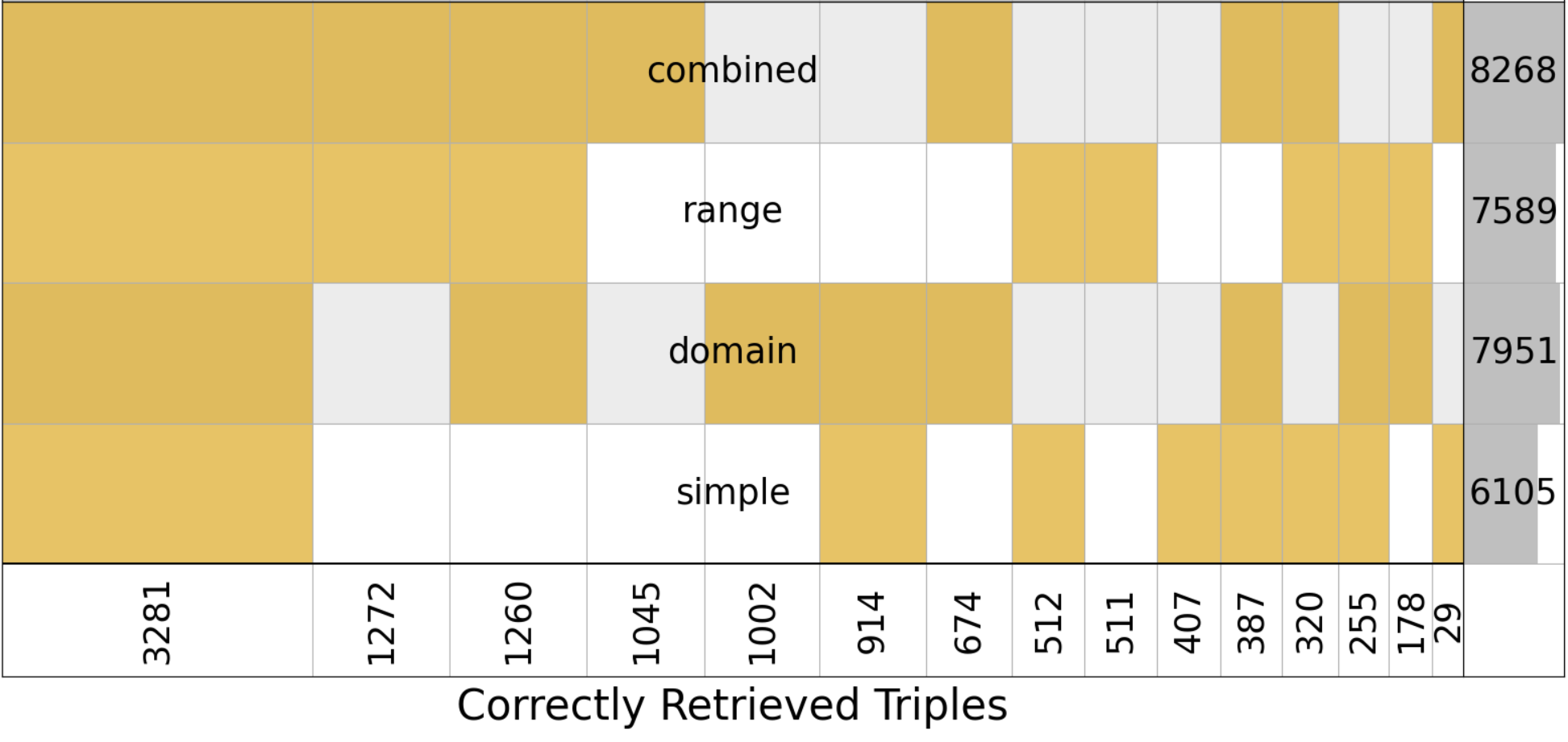}
         %\vspace{-2em}
         \caption{Appositive prompts}
         \label{fig:appositive_consistency}
     \end{subfigure}
    \hfill
    \vspace{-1em}
    \caption{Knowledge consistency for sInf added through (a) clausal and (b) appositive prompts for all intersections of correctly predicted triples by RoBERTa on the TREx corpus. 
    %These intersections are characterised by the number of agreeing sets (upper aggregation), agreeing sets per intersection (coloring in cell), size of each intersection (bottom number per column) and number of correctly predicted triples per prompt (aggregation per row). 
    %Overall we observe that RoBERTa is more consistent in the task of knowledge retrieval when given sInf through clausal prompts in comparison to the appositive prompts.Providing RoBERTa with sInf using clausal prompts achieves higher performance than appositive prompts.
    }
    \label{fig:experiment_three}
    \vspace{-1em}
\end{figure*}

\begin{figure}[t]
    \centering
    \includegraphics[width=\columnwidth]{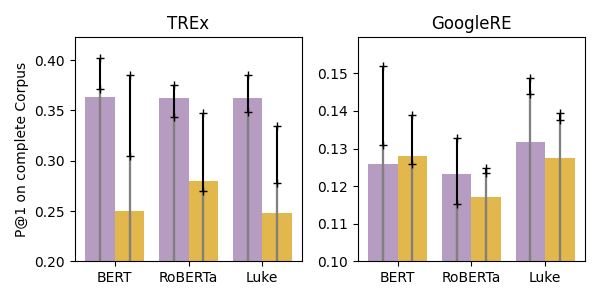}
    \vspace{-2em}
    \caption{Effect of combined domain+range information using either clausal (purple) or appositive (orange) syntax, compared to expected interval (black). Upper bound is choosing the better answer with either domain or range information, lower bound is choosing the one with the higher confidence.
}
    \label{fig:experiment_two}
        \vspace{-1em}
\end{figure}

\para{Impact of prompt syntax on efficient information combination (RQ2)} 
\label{ssec:Combinability}
We investigate if the performance differences between causal and appositive prompt syntax persist when we query with prompts that carry both range and domain information. 
Furthermore, we intend to assess if such prompts help the PLM to uncover synergies or act as noise. 
For this reason, we define a relative performance interval from the already observed \textit{rKR} performance on compound and complex prompts (analog for appositive). 
The low-end of performance is marked by choosing the answer with the highest confidence. 
The high-end of performance is estimated by choosing the answer to a prompt that signals the highest probability for the correct token. 
The results of this experiment are displayed in Figure \ref{fig:experiment_two}. 
A model capable of combining information given a prompt (purple or orange) would be between the bounds (black-part). 
If, given a prompt with both sInf, a model performs below the lower bound (grey-part), it is not able to recover the knowledge it previously displayed, inferring that the encoding of the prompt was noisy. 
As the PLM's response is determined by probability, exceeding the black part is only possible in highly unlikely data constellations. 
From Fig.~\ref{fig:experiment_two}, we can observe that appositive syntax is less often in the expected boundaries. 
Additionally, neither clausal nor appositive syntax enable all PLMs to reliably combine the information such that it performs above the lower bound (in the black-part). 
However, RoBERTa is able to combine information reliably for clausal syntax. 
In comparison, the behaviour is less reliable for appositive prompts. 
All models have a similar peak performance on the complete TREx corpus. 
These findings expand the conclusion of \citet{pandia2021sorting} that even potentially helpful information could be detrimental for \textit{rKR} performance.

\para{Impact of prompt syntax on knowledge consistency with regards to different levels of available information (RQ3)} 
\label{ssec:Consistency}
This analysis is only conducted on the TREx corpus as the sample size of correctly predicted samples given simple prompts is sufficiently large to support conclusive insights (GoogleRE: 277, ConceptNet: 513). 
We display the results as a multi-set Venn diagram\footnote{https://github.com/gecko984/supervenn} for RoBERTa in Figure \ref{fig:experiment_three} and report the numbers for the remaining models in the text. 
We can read the diagram from no sInf (bottom) to high sInf (top) content. 
Each row indicates the correctly predicted triples given a specific prompt/syntax type. 
Each column represents an intersection (subset) of triples known to different prompt types. 
Every intersection (column) has two properties: 
the size of the respective intersection (bottom row), and which prompt types intersect (coloring per cell, white means no intersection).
Lastly, the total cardinality (correct triples) for each prompt type is aggregated at the end of each row. 
Note that along the rows an upside-down staircase pattern should emerge indicating that less sInf offers worse performance and more sInf offers more information while retaining the already retrieved knowledge. 

We make the following observations:
(i)
Looking at the leftmost column of Fig.~\ref{fig:appositive_consistency}, for RoBERTa (R), only 54\%  (BERT (B): 51\%, Luke (L): 51\%) of the triples correctly retrieved by the simple prompt are also correctly retrieved by all three appositive prompts ($3281/6105 \approx 54\%$). We achieve much higher consistency with clausal prompts (R: $4891/6105 \approx 80\%$, B: 79\%, L: 78\%, Fig.~\ref{fig:clausal_consistency}). 
(ii) The number of knowledge triples retrieved only through the simple prompt is twice as big for appositive (R: 407, B: 476, L: 529) vs. clausal prompts (R: 194, B: 368, L: 256), implying that information is more distracting when added with appositive syntax. 
R recalls 86\% (B: 85\%, L: 86\%) of all triples known with the simple prompt through the compound-complex prompt. 
 The consistency is worse for the appositive syntax, where R recalls 62\% (B: 58\%, L: 62\%) with the combined appositive prompt, which equals a decrease in recall of 24\% compared to the clausal case (B: 27\%, L: 24\%). 
We can see that, in general, the appositive syntax performs worse and is less consistent when compared to clausal prompts. 
Our results agree with \citet{elazar2021measuring} that there are indeed substantial differences in consistency and performance between different paraphrases.

\begin{figure}[t]
    \centering
    \includegraphics[width=\columnwidth]{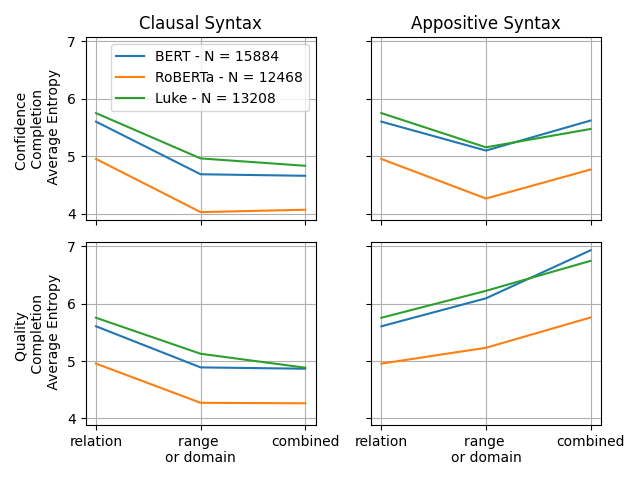}
    \vspace{-2em}
    \caption{Average binary entropy of response distribution of known subset (correct prediction in top 10) for the TREx corpora with differently completed prompts. Clausal syntax leaves less uncertainty. PLMs even generalize this loss in uncertainty to the combined setting given clausal syntax.}
    \label{fig:results_four}
    \vspace{-1em}
\end{figure}

\para{Impact of prompt syntax on response uncertainty (RQ4)} 
\label{ssec:Response uncertainty}
Here, we study the impact of syntax on the response uncertainty in PLMs by plotting the average binary entropy of the answer distributions with respect to the added information in Figure \ref{fig:results_four}. The binary entropy indicates the average bit-length needed to describe the answer set (i.e., entropy of 3 tells us that we effectively narrowed down the decision to 8 words). We follow \citet{gonen2022demystifying} who showed that information theoretic measures are a descriptor of prompt quality. 
We conduct our experiment on a subset containing the triples where the respective model retrieves the correct answer in the top 10 predictions for the simple prompt. 
Given this set is \textit{known} to the model we expect a decrease in uncertainty when sInf is added. 
The diagrams in the same column show the results for the prompts written in a respective syntax (clausal, appositive), whereas the diagrams in the same row show the results for a respective completion strategy. 

We observe that for clausal syntax the \textit{Quality Completion} offers uncertainty decrease with the addition of information, while for the appositive syntax, uncertainty increases as we add more information and this remains true for both completion strategies.

\section{Discussion}
\label{sec:discussion}

\citet{jiang2019know} showed that PLMs ability to retrieve information depends on the phrasing of the prompt. 
However, they make no qualitative statements about these phrases and even include non-natural language. 
In contrast, we classify different paraphrases based on their semantic and syntactic conditioning. 
We observe, in further detail, that adding supplementary domain and range information to simple prompts offers much less performance gain when realised via appositives than via clauses (RQ1).  
Adding the type-information in the sentence leaves the \textit{rKR} task mostly unchanged. 
Thus, in distinction to \citet{cao2021knowledgeable} and \citet{petroni2020context}, we can evaluate the in-sentence information processing. 
This reveals that appositive syntax adds noise in many cases, thereby lowering retrieval consistency (RQ3). 
Furthermore, a comparison of results on the combined prompts (RQ2) and the independent prompt (RQ1) shows that information that is helpful in isolation can be distracting when added in conjunction. 
Thus, we extend the findings from \citet{pandia2021sorting}  showing that it is not only misleading information that obfuscates the usability of a prompt. 
Lastly, we also show that the consistency of PLMs heavily depends on the syntactic relation between the prompts, which was only broached in \citet{elazar2021measuring}. 
The lower prevalence of appositives in the PLM training data as well as their semantic function of encoding conventional implications rather than assertions \citep{potts-2012-conventional}, might cause the lower performance for this syntax. 
Additionally, we observe that the worse-performing appositive prompts (those containing range information) tend to increase the dependency distance between the relation text and the masked object token, suggesting they perturb the models' ability to connect both. This further highlights the fragility of information flow in language representations achieved by PLMs \cite{ravichander2020systematicity}.
To counteract this fragility, a specialized training paradigm might be helpful, e.g. as proposed by \citet{elazar2021measuring}.

We argue that a knowledge-enhanced pre-training dataset as introduced by \citet{agarwal2020knowledge} would benefit from the integration of our findings. 
The introduction of multi-hop triples written in prompts that carefully follow a fitting syntax seems promising. 
We found that BERT is the model that performs on average the best, what is probably caused by the reliable data BERT is pre-trained on. 
However, peak performance for the TREx corpus of all models is at a similar level. This strengthens the already known fact that RoBERTa and its derivatives (i.e., Luke) learn more general representations of language in comparison to the older BERT model, 
as their performance gain comes with an increase in information. 
The findings presented in this paper might be used to add knowledge more reliably in approaches like K-BERT \cite{liu2020k}, KnowPrompt \cite{chen2022knowprompt}, or to aid knowledge graph construction as done in KG-BERT \cite{yao2019kg}.

\section{Conclusion \& Future Work}
\label{sec:conclusion}

In this paper, we introduced a controlled paraphrasing method and contributed the \probe{} to advance the investigation of knowledge retrieval from PLMs. Using \probe{}, we examined the impact of paraphrasing on the knowledge retrieval performance of PLMs, studying both clausal and appositive forms in conjunction with relation-specific sInf from Wikidata.

Our experimental findings reveal substantial variations in how PLMs process information based on prompt syntax. Particularly, we demonstrated that knowledge consistency is enhanced when prompts utilize clausal syntax. At the same time, we observe the vulnerability of language representations in PLMs, especially for appositive phrases. This susceptibility may be attributed to factors such as the prevalence of clausal syntax in the training data, the semantic function, and the syntactic interaction of words with the textual encoding of relations. This interpretation aligns with the conclusion by \citet{jiang2020can} that PLMs carry more knowledge than previously assumed and are highly sensitive to prompt paraphrasing. 
Although earlier research suggested limited benefits from adding shallow syntactical features, our study found that intentionally applied (clausal) syntax provides increased regularity that can be exploited by contextualized word representations. Therefore, we assume that harnessing these regularities through dedicated syntax-aware pre-training potentially facilitates a more robust knowledge representation.

To gain more conclusive insights into the knowledge retrieval capacities of PLMs, experiments on models trained on non-fictional and factually correct pre-training corpora are crucial. This approach can help distinguish between false pre-training knowledge and wrong retrieval. 
Exclusively pre-trained PLMs, on non-fictional, peer-reviewed corpora like Wikipedia or scholarly publications, or synthetic corpora \cite{agarwal2020knowledge}, can provide promising insights.

For future work, we plan to expand the template to include more diversity in syntax and knowledge. This involves applying our meta-template to additional relations derived from knowledge bases such as Wikidata and incorporating more complex syntactical structures. 

\section{Limitations}
All our investigations are done on English text and on one token objects. However, the LAMA-probe has already inspired multilingual knowledge retrieval \cite{kassner2021multilingual}, as well as multi-token knowledge retrieval \cite{kalo2022kamel}, which are out of the scope of this work.
Although our experimental set-up works on a variety of relations, models, and prompt typologies, we have only considered base models.
An additional testing of larger models like the `large' alternatives of BERT \cite{devlin2018bert} and RoBERTa \cite{liu2019roberta} would provide further insights.
Moreover, we only investigated the addition of type information per entity (subject, object) of a knowledge triple. Another variation to our used information could be the use of different types of sInf (e.g., `Obama is born in 1961 and was born in Hawaii.').
Additionally, one could extend our research to test the addition of more entity-specific information (i.e., `Obama is a president and was born in 1961 and was born in Hawaii.'). 

\section{Ethical Considerations}
Our research is not using any personal data and has no direct ethical implications. However, applying the proposed approach to retrieve knowledge from PLMs might reproduce societal biases encoded in the models (e.g., retrieval performance for male scientists might be higher than for female scientists).  
Additionally, we strive for the lowest energy footprint by working with the base-types configurations of already pre-trained PLMs. 

\bibliography{sample-base}

\appendix

\end{document}